\documentclass{article}

     \usepackage[preprint, nonatbib]{neurips_2020}

\usepackage[utf8]{inputenc} 
\usepackage[T1]{fontenc}    
\usepackage{hyperref}       
\usepackage{url}            
\usepackage{booktabs}       
\usepackage{amsfonts}       
\usepackage{nicefrac}       
\usepackage{microtype}      
\usepackage{comment}
\usepackage{graphicx} 
\usepackage{pdfpages}

\title{Does Explainable Artificial Intelligence Improve Human Decision-Making?}

\author{%
  Yasmeen Alufaisan \\
  EXPEC Computer Center \\
  Saudi Aramco \\
  \texttt{yasmeen.alufaisan@aramco.com} \\
   \And
   Laura R.~Marusich \\
   U.S. Army Combat Capabilities Development Command Army Research Laboratory South\\
   at the University of Texas at Arlington\\
  \texttt{laura.m.cooper20.civ@mail.mil} \\
   \AND
   Jonathan Z.~Bakdash \\
   U.S. Army Combat Capabilities Development Command Army Research Laboratory South\\
   at the University of Texas at Dallas\\
  \texttt{jonathan.z.bakdash.civ@mail.mil} \\
   \And
   Yan Zhou \\
   Department of Computer Science \\
   University of Texas at Dallas \\
   \texttt{zhou07@gmail.com} \\
   \And
   Murat Kantarcioglu \\
   Department of Computer Science \\
   University of Texas at Dallas \\
  \texttt{muratk@utdallas.edu} \\
}
\begin{document}

\maketitle

\begin{abstract}
Explainable AI provides insight into the \emph{why} for model predictions, offering potential for users to better understand and trust a model, and to recognize and correct AI predictions that are incorrect. Prior research on human and explainable AI interactions has focused on measures such as interpretability, trust, and usability of the explanation. Whether explainable AI can improve actual human decision-making and the ability to identify the problems with the underlying model are open questions. Using real datasets, we compare and evaluate objective human decision accuracy without AI (control), with an AI prediction (no explanation), and AI prediction with explanation. We find providing any kind of AI prediction tends to improve user decision accuracy, but no conclusive evidence that explainable AI has a meaningful impact.  Moreover, we observed the strongest predictor for human decision accuracy was AI accuracy and that users were somewhat able to detect when the AI was correct versus incorrect, but this was not significantly affected by including an explanation. Our results indicate that, at least in some situations, the \emph{why} information provided in explainable AI may not enhance user decision-making, and further research may be needed to understand how to integrate explainable AI into real systems.   
\end{abstract}

\section{Introduction}\label{intro}
Explainable AI is touted as the key for users to ``understand, appropriately trust, and effectively manage [AI systems]''~\cite{gunningExplainableArtificialIntelligence2017} with parallel goals of achieving fairness, accountability, and transparency~\cite{sokol2019fairness}. There are a multitude of reasons for explainable AI, but there is little empirical research for its impact on human decision-making ~\cite{miller2019explanation, adadi2018peeking}. 
Prior behavioral research on explainable AI has primarily focused on human understanding/interpretability, trust, and usability for different types of explanations~\cite{doshi2017towards, hoffman2018metrics, ribeiro2016should, ribeiro2018anchors, lage2019evaluation}. 


To fully achieve fairness and accountability, explainable AI should lead to better human decisions. Earlier research demonstrated that explainable AI can be understood by people~\cite{ribeiro2018anchors}. Ideally, the combination of humans and machines will perform better than either alone~\cite{adadi2018peeking}, such as computer-assisted chess~\cite{cummings2014man}, but this combination may not necessarily improve the overall accuracy of AI systems. While (causal) explanation and prediction share commonalities, they are not interchangeable concepts ~\cite{adadi2018peeking, shmueli2010explain, edwards2018enslaving}. Consequently, a "good" explanation, interpretable model predictions, may not be sufficient for improving actual human decisions~\cite{adadi2018peeking, miller2019explanation} because of heuristics and biases in human decision-making rather than optimality
~\cite{kahneman2011thinking}. Therefore, it is important to demonstrate whether, and what types of, explainable AI can improve the decision-making performance of humans using that AI, relative to performance using the predictions of "black box" AI predictions with no explanations and also compared to no AI prediction.

In this work, we empirically investigate whether explainable AI improves human decision-making using a two-choice classification experiment with real-world data. Using human subject experiments, we compared three different settings where a user needs to make a decision: 1) No AI prediction (Control), 2) AI predictions but no explanation, and 3) AI predictions with explanations. Our results indicate that, while providing the AI predictions tends to help users, the \emph{why} information provided in explainable AI does not specifically enhance user decision-making. 

\section{Background and Related Work}
Using Doshi-Velez and Kim's ~\cite{doshi2017towards} framework for interpretable machine learning, our current work focuses on real humans, simplified tasks. 
Because our objective is on evaluating decision-making, we do not compare different types of explanations and instead used one of the best available explanations: anchor LIME~\cite{ribeiro2018anchors}. We use real tasks here, although our tasks involve relatively simple decisions with two possible choices. Additionally, we use lay individuals rather than experts.  Below, we discuss prior work that is related to our experimental approach.

\subsection{Explainable AI/Machine Learning}
While machine learning models largely remain opaque and their decisions are difficult to explain, there is an urgent need for machine learning systems that can ``explain'' their reasoning. For example, European Union regulation requires a ``right to explanation'' for any algorithms that make decisions significantly impacting users with user-level predictors~\cite{euro_reg}. In response to the lack of consensus on the definition and evaluation of interpretability in machine learning, Doshi-Velez and Kim~\cite{doshi2017towards} propose a taxonomy for the evaluation of interpretability focusing on the synergy among human, application, and functionality. They contrast interpretability with reliability and fairness, and discuss scenarios in which interpretability is needed~\cite{46160}. To unmask the incomprehensible reasoning made by these machine learning/AI models, researchers developed explainable models that are built on top of the machine learning model to explain their decisions. There are two categories of explainable models that provide explanations for the decisions made by machine learning models: feature-based and rule-based models. The feature-based models resemble feature selection where the model outputs the top features that explain the machine learning prediction and their associated weights~\cite{QII,ribeiro2016should}. The rule-based models provide simple if-then-else rules to explain predictions~\cite{ribeiro2018anchors,IT:2017}. It has been proven that rule-based models provide higher precision when compared to feature-based models~\cite{ribeiro2018anchors, IT:2017}.   

Lou et al.~\cite{lou2012intelligible} investigate the performance of generalized additive models (GAMs) that combine single-feature models through a linear function. Each single-feature model is referred to as a shape function that can be arbitrarily complex; however, there are no interactions between features. Therefore, GAMs are  more accurate than simple linear models and can be easily interpreted by users. They propose  a new model based on tree ensembles with an adaptive number of leaves. They also investigate how  different  shape functions  influence  the  additive model.  Their empirical study suggests that shallow bagged trees with gradient boosting appear to be the best method on low to medium dimensional datasets. Anchor LIME is an example of the current state-of-the-art explainable rule-based model~\cite{ribeiro2018anchors}. It is a model-agnostic system that can explain predictions generated by any machine learning model with high precision. The model provides rules, referred to as anchors, to explain the prediction for each instance. A rule is an anchor if it sufficiently explains the prediction locally such that any changes to the rest of the features, features not included in the anchor, do not affect the prediction. Anchors can be found in two different approaches: bottom-up approach and beam search. More details of anchor LIME can be found in~\cite{ribeiro2018anchors}. Wang et al.~\cite{JMLR:v18:16-003} present a machine learning algorithm that produces Bayesian rule sets (BRS) comprising short rules in the disjunctive normal form. They develop two probabilistic models with prior parameters that allow the user to specify a desired size and shape and  balance between accuracy and interpretability. They apply two priors---beta-binomials and Poisson distribution---to constrain the  rule generation process and provide theoretical bounds for reducing computation by iteratively pruning the search space. In our experiments,  we use anchor LIME to provide explanations for all our experimental evaluations due to the high human precision of anchor LIME as reported in~\cite{ribeiro2018anchors}.  

\subsection{Human Experiments with Explainable AI and Human Decision-Making}
Prior human experiments with explainable AI have concentrated on interpretability, trust, and subjective measures of usability (preferences and satisfaction),  with work on decision-making performance remaining somewhat limited. For example, one study used a classification task with text information and varied the complexity of explanations. As the explanation complexity increased, users demonstrated more difficulty with interpretability (based on slower responses) and decreased usability, but no statistically significant impact on decision accuracy ~\cite{lage2019evaluation}. Comparisons were only among explanations of varying complexity; there was no AI prediction without an explanation. For tasks with text and images, the addition of explanations increased trust in the classifier and also helped users select the classifier with higher accuracy~\cite{ribeiro2016should}. Another study was primarily designed to compare human performance between two different types of explanations and did include baseline conditions with no explanation ~\cite{ribeiro2018anchors}. In this work, users with anchors were more accurate and faster to predict out of sample model classification and strongly preferred the anchors~\cite{ribeiro2018anchors}. 
Earlier results suggest explainable AI is interpretable, trustworthy, and usable to varying degrees, 
but 
this is not necessarily the same as a person making real-world decisions about the underlying data, such as whether to actually use the AI's prediction, whether the AI has made an error, and the role of explanations.


Psychological research on decision-making suggests that "good" interpretability may not by itself be sufficient for enhancing decisions. People are not necessarily rational (i.e., maximizing an expected utility function). Instead, decisions are often driven by heuristics and biases ~\cite{kahneman2011thinking}. Also, providing more information, even if relevant, does not necessarily lead people to making better decisions~\cite{gigerenzer2009h}. Bounded rationality in human decision-making using satisfying (near optimality) ~\cite{gigerenzer2009h} is an alternative theory to heuristics and biases~\cite{kahneman2011thinking}. Regardless of the theoretical account for human decision-making, people, which can include experts ~\cite{dawes1989clinical}, generally do not make fully optimal decisions.

At a minimum, explainable AI should not be detrimental to human decision-making. The literature on decision aids (a computational recommendation or prediction, typically without an explicit explanation) has mixed findings for human performance. Sometimes these aids are beneficial for human decision-making, whereas at other times they have negative effects on decisions~\cite{kleinmuntz1993information, skitka1999does}. These mixed findings may be attributable to the absence of explanations. A common reason for explanation is to improve predictions or decisions~\cite{keil2006explanation}.  


\section{Methods}\label{methods}
In this section, we first describe the two datasets used in our experiments. We then provide the details of our experimental design and hypotheses, participant recruitment, and general demographics of our sample. 

\subsection{Dataset}
To conduct our experiments, we choose two different datasets that have been heavily used in prior research that tries to understand algorithmic fairness and accountability issues. 
For example, the COMPAS dataset has been used to detect potential biases in the criminal justice system~\cite{COMPAS}. The Census income dataset, which has been used to test many machine learning techniques, involves predictions of individuals' income status. This has been associated with potential biases in making decisions such as access to credit and job opportunities. 

We choose these datasets primarily because they both involve real-world contexts that are understandable and engaging for human participants. Further, the two datasets differ widely in the number of features and in the overall accuracy classifiers can achieve in their predictions. This allows us to explore the effects of these differences on human performance; in addition, it ensures that our findings are not limited only to a specific dataset. We briefly discuss each dataset in more detail below.

\textbf{COMPAS} stands for Correctional Offender Management Profiling for Alternative Sanctions~\cite{COMPAS}. It is a scoring system used to assign risk scores to criminal defendants to determine their likelihood of becoming a recidivist. 
The data has 6,479 instances and 7 features. These features are gender, age, race, priors count, and charge degree risk score, and whether the defendants re-o?ended in two years or not. We let the binary re-o?ending feature be our class. 

\textbf{Census income (CI)} data contains information used to predict individuals' income~\cite{Lichman:2013}. It has 32,561 instances and 12 features. These features are age, work class, education, marital status, occupation, relationship, race, sex, capital gain, capital loss, hours per week, and country. The class value is low income (less or equal to 50K) or high income (greater than 50K). We preprocessed the dataset to allow equal class distribution \footnote{The CI dataset is from 1994. We adjusted for inflation by using a present value of 88K. From 1994 to January 2020 (when the experiment was run) inflation in the United States was 76.45\%: \url{https://www.wolframalpha.com/input/?i=inflation+from+1994+to+jan+2020}}. 

\subsection{Experimental Design}\label{design}
Prior results demonstrate that people interpret, trust, and prefer explainable AI, suggesting it will improve the accuracy of human decisions. Hence, our primary hypotheses are that explainable AI would aid human decision-making. The hypotheses (H.) are as follows:  
\begin{enumerate}
    \item[H.1]  Explainable AI enhances the decision-making  process compared to only an AI prediction (without explanation) and a control condition with no AI.
    \subitem[H.1.a] A participant performs above chance in prediction tasks.
    \item[H.2] A participant's decision accuracy is positively associated with AI accuracy.
    \item[H.3] Average participant's decision accuracy does not outperform AI accuracy. 
    \item[H.4] Participants outperform AI accuracy more often with explainable AI over AI prediction alone. 
    \item[H.5] Participants follow explainable AI recommendation more often than AI-only recommendation.
    \item[H.6] Explainable AI increases a participant's decision confidence. 
    \item[H.7] A participant's decision confidence is positively correlated with the accuracy of his/her decision. 
\end{enumerate}

To investigate these hypotheses, we used a 2 (Dataset: Census and COMPAS) x 3 (AI condition: Control, AI, and AI with Explanation) between-participants experimental design. The three AI conditions were:
\begin{itemize}
    \item Control: Participants were provided with no prediction or information from the AI.
    \item AI: Participants were provided with only an AI prediction.
    \item AI with Explanation: Participants received an AI prediction, as well as an explanation of the prediction using anchor LIME~\cite{ribeiro2018anchors}. 
\end{itemize}

\begin{figure}[h]
    \centering
    \includegraphics[width=0.85\linewidth]{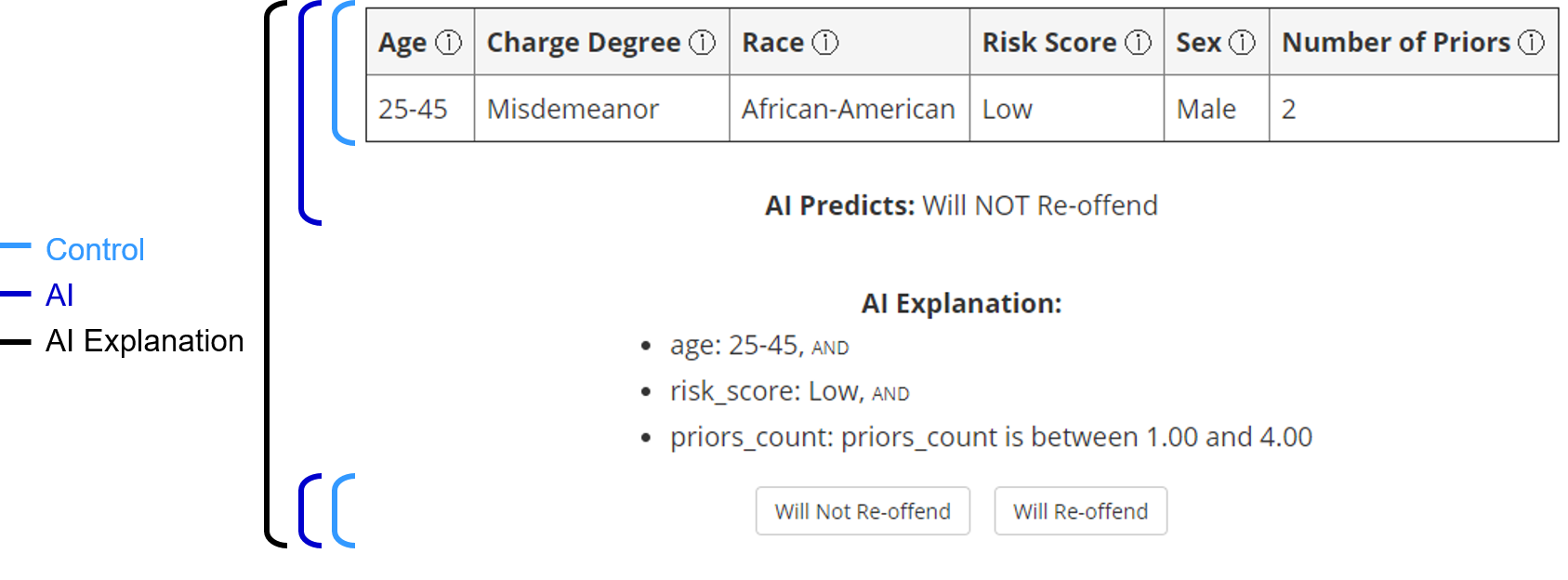}
    \caption{
        Example from the study demonstrating the information appearing in the three AI conditions for a trial from the COMPAS dataset condition.
        \label{fig:screenshot}
    }
\end{figure}
To achieve more than 80\% statistical power to detect a medium effect size for this design, we planned for a sample size of ${N}$ = 300 (50 per condition). 
    
In all conditions, each trial consists of a description of an individual and a two-alternative forced choice for the classification of that individual. Each choice was correct on 50\% of the trials, thus chance performance for human decision-making accuracy was 50\%. Additionally, an AI prediction and/or explanation may appear, depending on the AI condition (see Figure \ref{fig:screenshot}). After a decision is made, participants are asked to enter their confidence in that choice, on a Likert scale of 1 (No Confidence) to 5 (Full Confidence). Feedback is then displayed, indicating whether or not the previous choice was correct.

We compared the prediction accuracy of Logistic Regression, Multi-layer Perceptron Neural Network with two layers of 50 units each, Random Forest, Support Vector Machine (SVM) with rbf kernel, and selected the best classifier for each dataset. 
We chose a Multi-layer Perceptron Neural Network for Census income data where it resulted in an overall accuracy of 82\% and SVM with rbf kernel for COMPAS data with an overall accuracy of 68\%. Census income accuracy closely matches the accuracy reported in the literature~\cite{Lichman:2013,Kaggle_Census,CIC16} and COMPAS accuracy matches the results published by ProPublica~\cite{COMPAS}. We split the data to 60\% for training and 40\% for testing to allow enough instances for the explanations generated using anchor LIME~\cite{ribeiro2018anchors}.

In our behavioral experiment, 50 instances were randomly sampled without replacement for each participant. Thus, AI accuracy was experimentally manipulated for participants (Census: mean AI accuracy = 83.85\%, sd = 3.67\%; COMPAS: mean AI accuracy = 69.18\%, sd = 4.65\%). Because of the sample size and large number of repeated trials per participant, there was no meaningful difference in mean AI accuracy for participants in the AI condition versus those in the AI explanation condition (${p} = 0.90$).  

\subsection{Participant Recruitment and Procedure}

The experiment was created in jsPsych~\cite{de2015jspsych} and hosted on the Volunteer Science platform~\cite{radford2016volunteer} \footnote{\url{https://volunteerscience.com/}}. Participants were recruited using Amazon Mechanical Turk (AMT) and were compensated \$4.00 each. We collected data from 50 participants in each of the six experimental conditions, for a total of 300 participants (57.67\% male). Most participants were 18 to 44 years old (80.67\%). The behavioral experiment research was approved as exempt by our institutional review board. 
 


Participants read and agreed to a consent form, then received instructions on the task, specific to the experimental condition they were assigned to. They completed 10 practice trials, followed by 50 test trials and a brief questionnaire assessing general demographic information and comments on strategies used during the task. The median time to complete the practice and test trials was 18 minutes.

\section{Results and Discussion}\label{results}
In this section we analyze and describe the effects of dataset, AI condition, and AI accuracy on the participants' decision-making accuracy, ability to outperform the AI, adherence to AI recommendations, confidence ratings, and reaction time. 

\subsection{Participant Decision-Making Accuracy}
\begin{figure}[!htb]
    \centering
    \begin{minipage}{.45\textwidth}
    \centering
    \includegraphics[trim={0.25cm 0.25cm 0.25cm 0.25cm}, clip, width=0.9\textwidth]{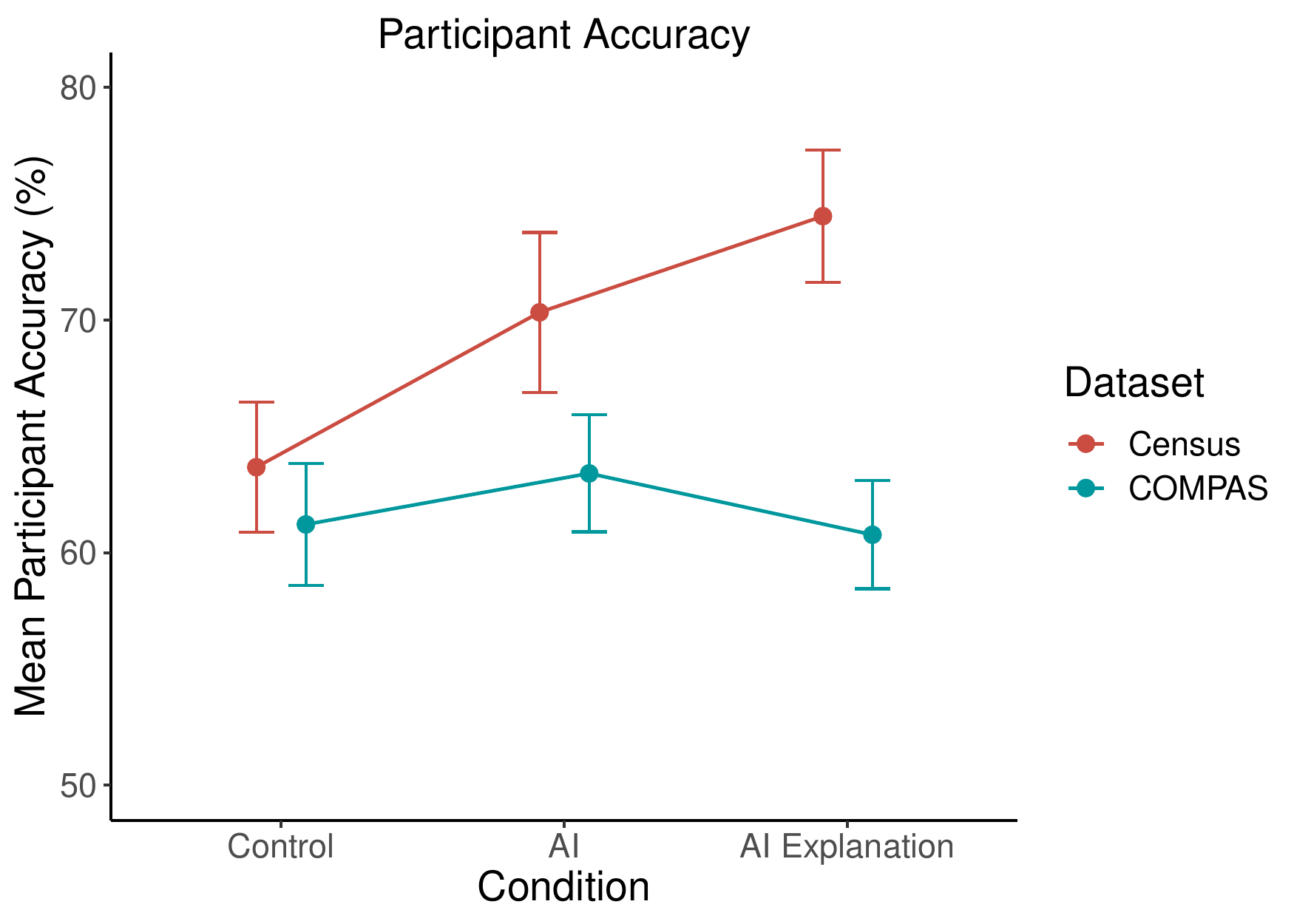}
    \caption{
        Mean participant accuracy in each AI and dataset condition. Error bars represent 95\% confidence intervals.
        \label{fig:accuracyplot}
    }
    \end{minipage}
    \hfill
    \begin{minipage}{.45\textwidth}
    \centering
    \includegraphics[width=.9\textwidth]{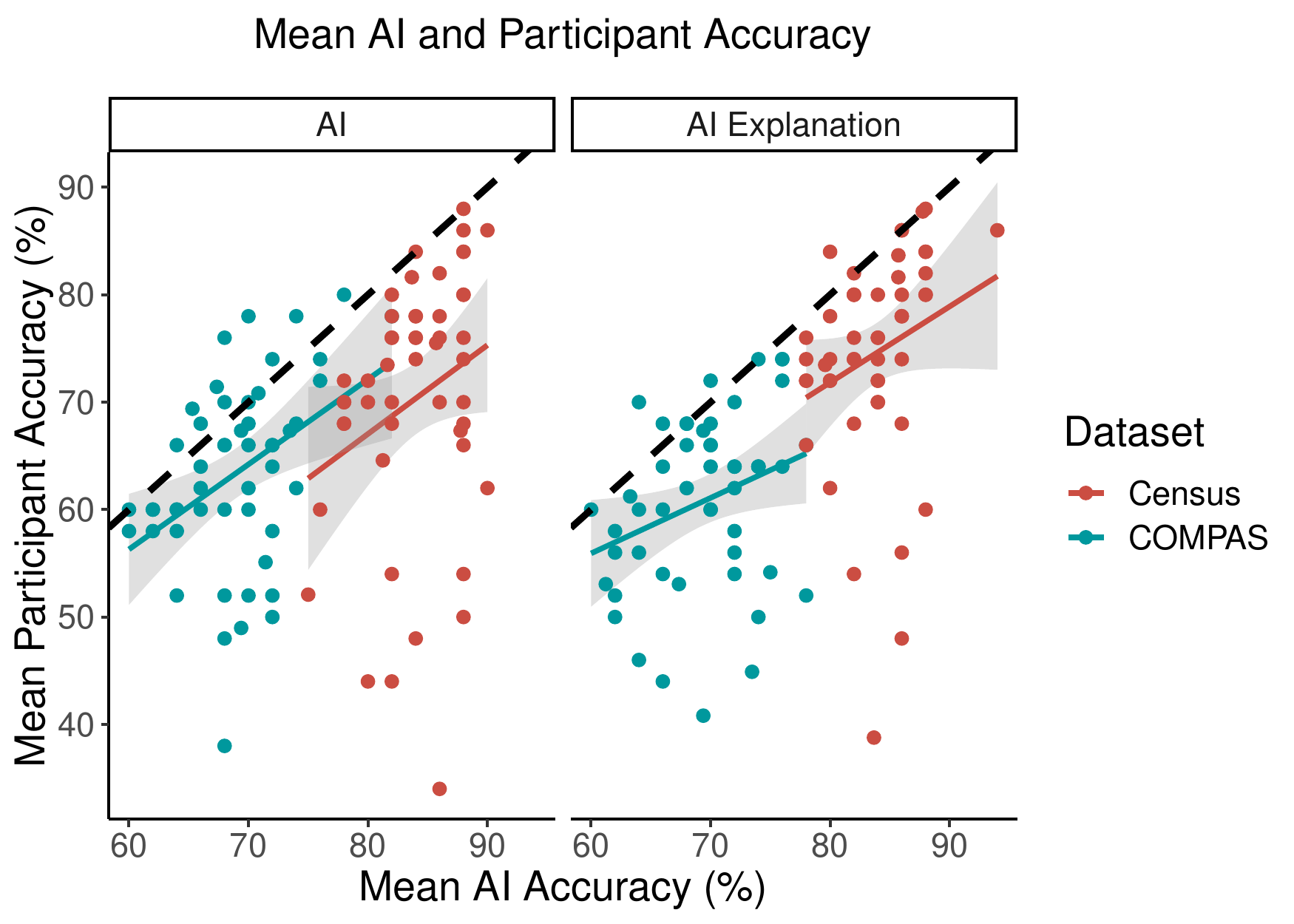}
    \caption{
        Mean AI accuracy (per participant) and mean participant accuracy by AI and AI Explanation and the two datasets. The shaded areas represent 95\% confidence intervals. 
        \label{fig:AIdmacc}}
    \end{minipage}
\end{figure}

We compared participants' mean accuracy in the experiment across conditions using a 2 (Dataset) x 3 (AI) factorial Analysis of Variance (ANOVA)  (see Figure \ref{fig:accuracyplot}). We found significant main effects, with a small effect size for AI condition ($F(2,294) = 8.19, {p} < 0.001, \eta^2 = 0.04$) and a nearly large effect for dataset condition ($F(1,294) = 46.51, {p} < 0.001, \eta^2 = 0.12$). In addition, there was a significant interaction with a small effect size ($F(2,294) = 8.38, {p} < 0.001, \eta^2 = 0.05$), indicating that the effect of AI condition depended on the dataset. Specifically, the large effect for increased accuracy with AI was driven by the Census dataset. 

Contrary to H.1, explainable AI did not substantially improve decision-making accuracy over AI alone. We followed up on significant ANOVA effects by performing pairwise comparisons using Tukey's Honestly Significant Difference. These post-hoc tests indicated that participants who viewed the Census dataset showed improved accuracy over control when given an AI prediction (${p} < 0.01$) and higher accuracy with AI explanation versus control (${p} < 0.001$), but there was no statistically significant difference in participant accuracy for AI compared to AI explanation (${p} = 0.28$). However, the COMPAS dataset had no significant differences in participant accuracy across pairwise comparisons for the three AI conditions (${ps} > 0.75$). 
Also, the mean participant accuracy for the COMPAS control condition (mean = 63.7\%, sd = 9.24\%) was comparable to participant accuracy for prior decision-making research using the same dataset (mean = 62.8\%, sd = 4.8\%)~\cite{dressel2018accuracy}. 

There was strong evidence supporting H.1.a; the vast majority of participants had mean accuracy exceeding guessing (50\% accuracy). The overall participant accuracy across all conditions was 65.65\% (sd = 10.92\%), with 90\% (or 270 out of 300) of participants performing above chance on the classification task. This indicates that the task was challenging but feasible for almost all participants. 

\subsection{AI Accuracy and Participant Decision-Making Accuracy}
We also evaluated the effect of the randomly varied AI accuracy 
for each participant on their decision-making accuracy. We used linear regression to analyze this relationship, specifying participant accuracy as the dependent variable and the following as independent variables: mean AI accuracy (per participant), AI condition, and dataset condition, see Figure \ref{fig:AIdmacc}. Regressions are represented by the solid lines with the shaded areas representing 95\% confidence intervals. The control condition is not included in the analysis or figure, because the accuracy of the AI is not relevant if no AI prediction is presented to the participant. The overall regression model was significant with a large effect size, $F(4, 195) = 21.23, {p} < 0.001, 
R_{adjusted}^2 = 0.29$. Consistent with H.2, there was a large main effect for AI accuracy ($\beta = 0.70, {p} < 0.001, R^2 = 0.28$). Also, there was a small AI accuracy and dataset interaction ($\beta = -0.07, {p} < 0.01, R^2 = 0.03)$, reflecting the same interaction depicted in Figure \ref{fig:accuracyplot}. There were no significant regression differences for the dataset or AI versus AI Explanation, ${ps} > 0.60$; there was no significant effect of dataset because it largely drove AI accuracy. (It is not just that participants perform better with the higher mean AI accuracy of the Census dataset; both datasets had large positive relationships with participant accuracy and corresponding mean AI accuracy shown in Figure \ref{fig:AIdmacc}.) 


\subsubsection{Outperforming the AI Accuracy}

An interesting question is whether the combination of AI and human decision-making can outperform either alone. The previous analyses showed that the addition of AI prediction information improved human performance over controls with humans alone. We also evaluated how often the human decision-making accuracy outperformed the accuracy of the corresponding mean AI prediction accuracy, which was experimentally manipulated. Although most participants performed well above chance, only a relatively small number of participants had decision accuracy exceeding their mean AI prediction (7\% or 14 out of 200). This result largely supports H.3 and also shown above in Figure \ref{fig:AIdmacc} where each dot represents an individual; also shown above, dots above the black dashed line show the participants that outperformed their mean AI prediction. The black dashed line shows equivalent performance for mean AI accuracy and mean participant accuracy. 
\begin{table}[h]
\centering
\caption{Number of participants with decision accuracy exceeding their mean AI accuracy.}
\label{tab:outperAI} 
 \begin{tabular}{|c c c|} 
 \hline
 Dataset/Condition & AI & AI Explanation \\ 
 \hline\hline
 Census & 0 & 1 \\ 
 \hline
 COMPAS & 10 & 3 \\
 \hline
\end{tabular}
\end{table}

\subsubsection{Adherence to AI Model Predictions}





\begin{figure}[h]
    \centering
    \includegraphics[width = .35\linewidth]{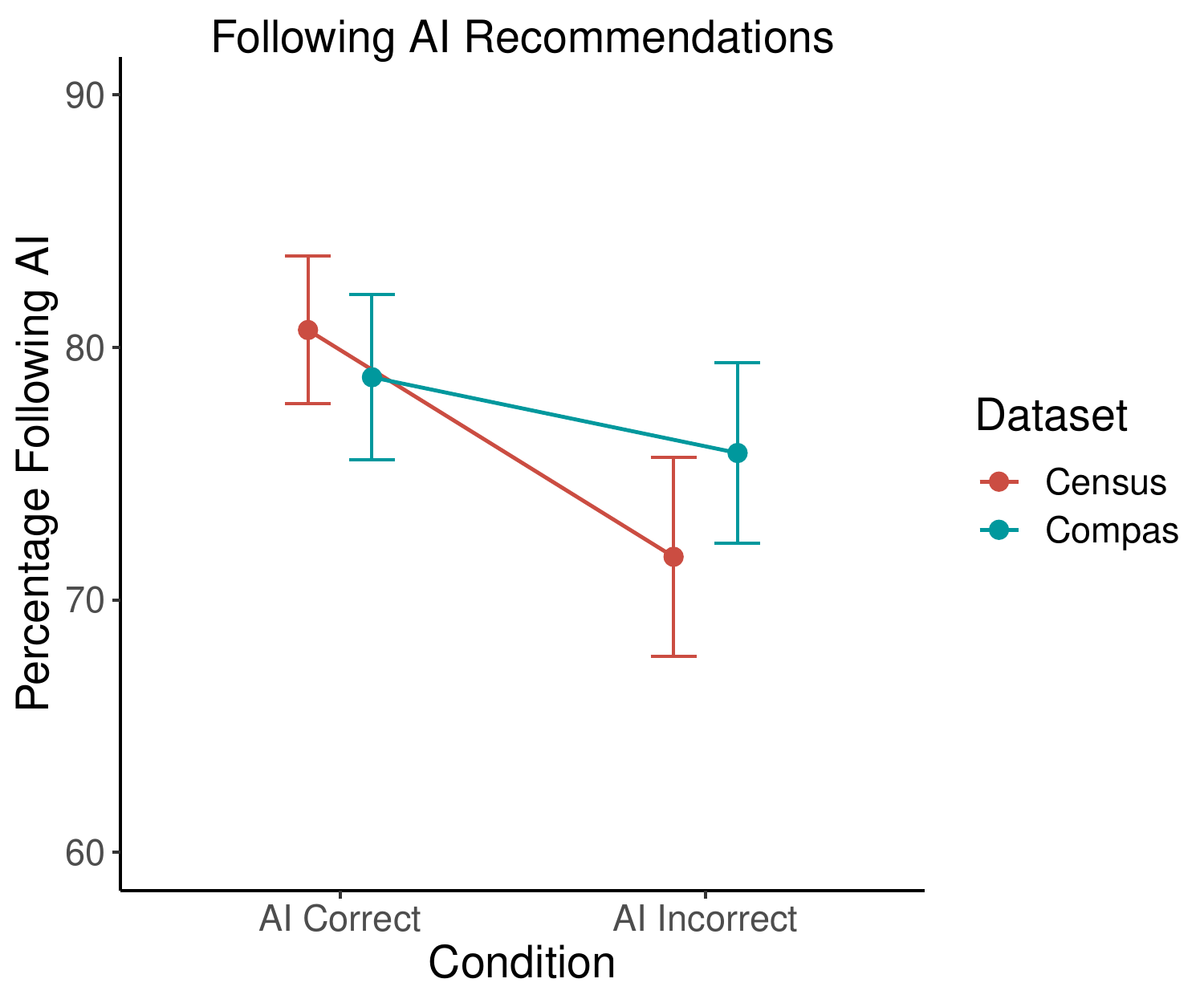}
    \caption{
        Mean proportion of participant choices matching AI prediction as a function of whether the AI correct or incorrect and the dataset condition. Error bars represent 95\% confidence intervals. To simplify this figure, results were collapsed for the AI and AI Explanation conditions, which did not have a significant main effect, ${p} = 0.62$.  
        \label{fig:aiplot2}
    }
\end{figure}

Participants followed the AI predictions more often when the AI was correct versus when the AI was incorrect, indicating some recognition of when the AI makes bad predictions (see Figure \ref{fig:aiplot2}, $F(1,196) = 36.15, {p} < 0.001, \eta_p^2 = 0.16$). This was consistent with participant sensitivity to AI recommendations, evidence for H.2. Also, participants were better able to recognize correct versus incorrect AI predictions when they were in the Census condition, demonstrated in the significant interaction between AI correctness and dataset, $F(1,196) = 9.01, {p} < 0.01, \eta_p^2 = 0.04$.  None of the remaining ANOVA results were significant, ${ps} > 0.16$. Thus, there was no evidence for higher adherence to recommendations with explainable AI, which rejected H.5. 

\subsection{Confidence Ratings}
\begin{figure}[h]
    \centering
    \begin{minipage}{0.45\textwidth}
    \centering
    \includegraphics[width =\linewidth]{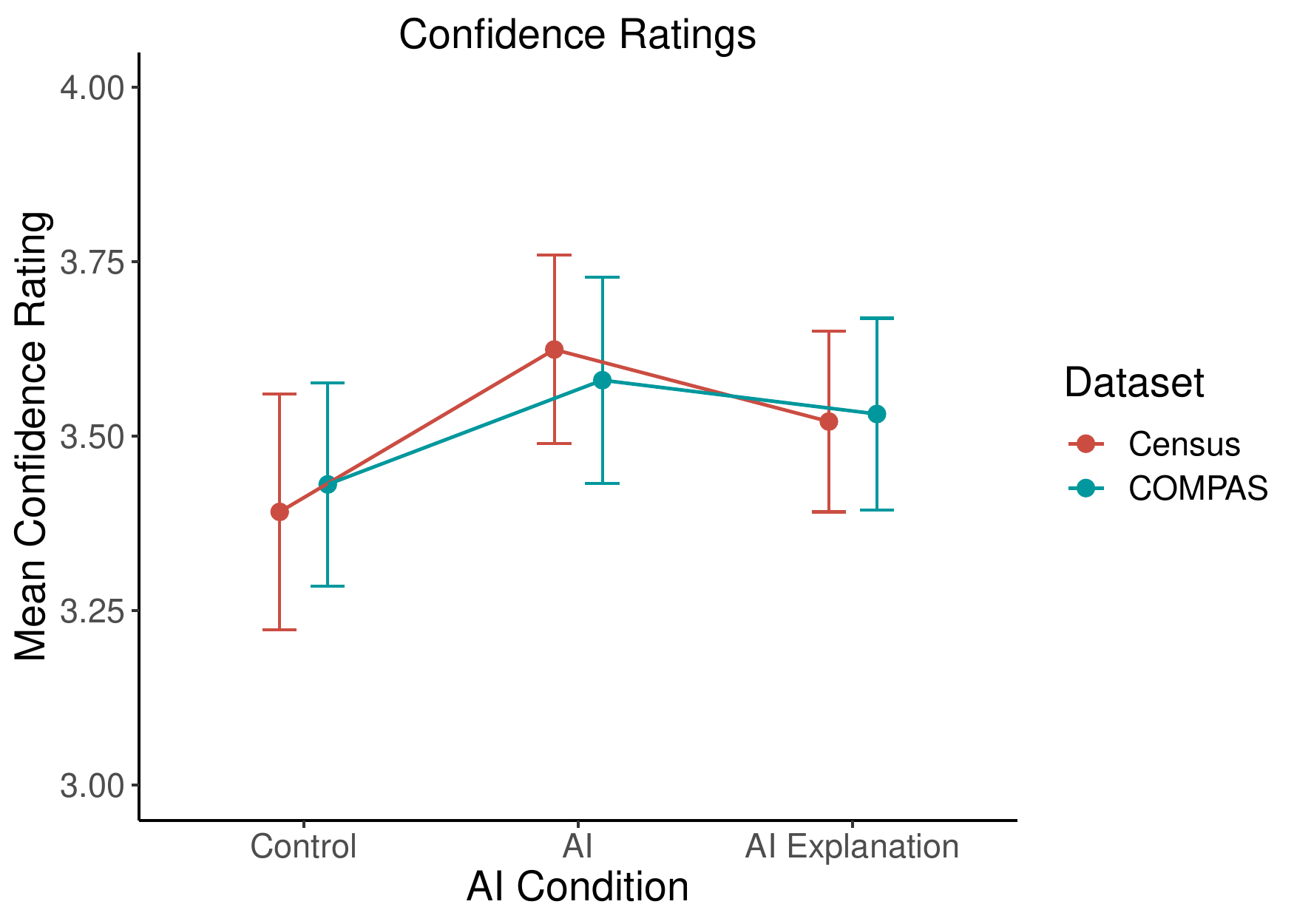}
    \caption{
        Mean participant confidence ratings in each AI condition. Error bars represent 95\% confidence intervals.
        \label{fig:confplot}
    }
    \end{minipage}%
    \hfill
    \begin{minipage}{0.45\textwidth}
    \centering
        \includegraphics[width =0.95\linewidth]{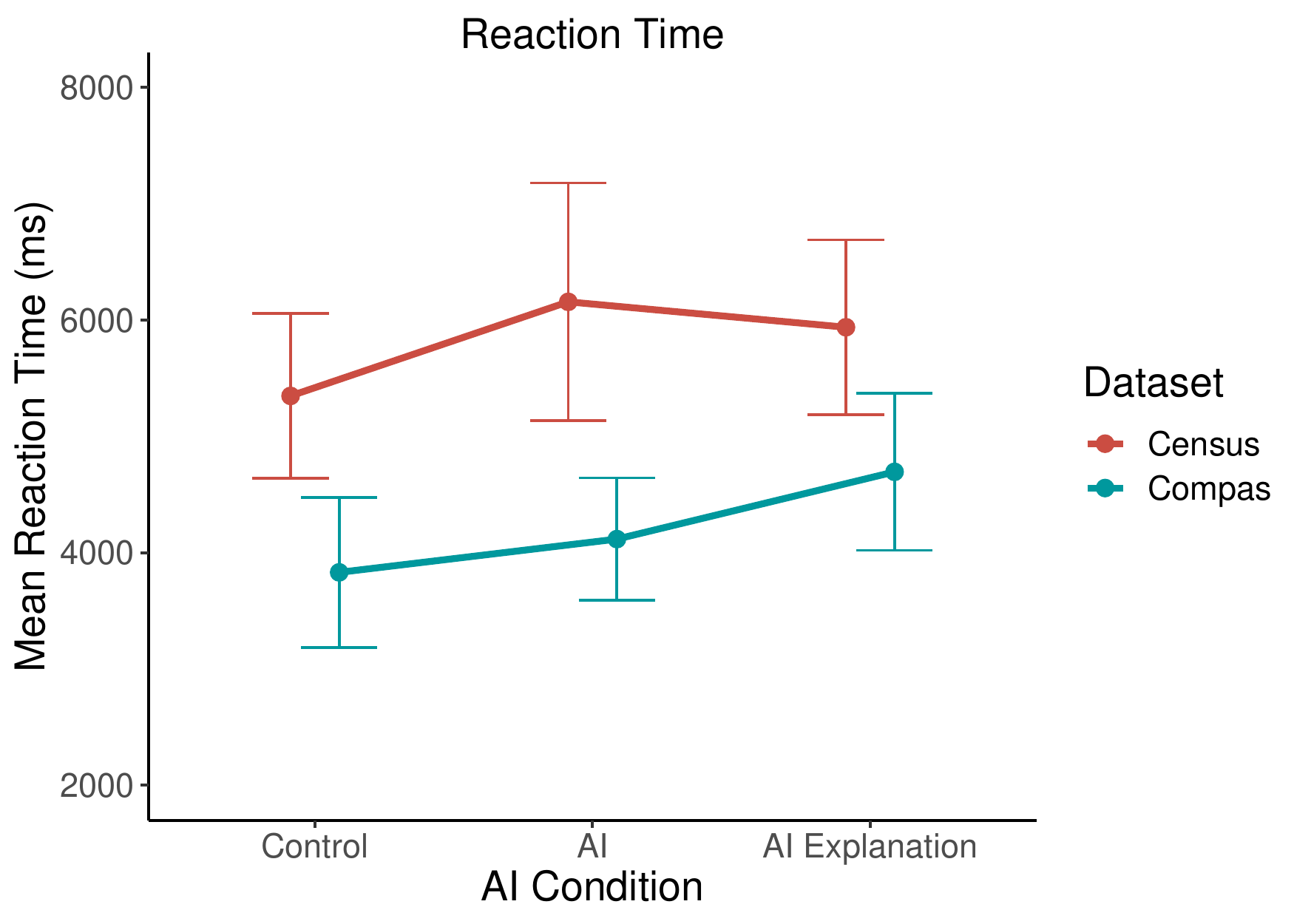}
    \caption{
        Mean reaction time in each AI and dataset condition. Error bars represent 95\% confidence intervals.
        \label{fig:rtplot}
    }
    \end{minipage}
\end{figure}

We found that AI (without and with explanation) resulted in slightly increased mean confidence.  There was a small effect of AI condition on mean confidence (see Figure \ref{fig:confplot}, $F(2,294) = 3.58, {p} = 0.03, \eta^2 = 0.02$). Post hoc tests indicated participants had significantly lower mean confidence in the control condition than AI, ${p} < 0.03$, but there were no statistical differences for other pairwise comparisons, ${ps} > 0.25$. This contradicted H.6, and there was no evidence of a confidence increase with explanations. In addition, there was no evidence for a main effect of dataset condition or interaction, ${ps} > 0.84$. 

Confirming H.7, we found a positive relationship for accuracy and confidence rating within individuals indicating that participants' confidence ratings were fairly well-calibrated with their actual decision accuracy. We calculated each participant's mean accuracy at each confidence rating they used, and then conducted a repeated measures correlation~\cite{bakdash2017repeated} ($r_{rm} = 0.48, p < 0.001$). 

\subsection{Additional Results}
We also assessed reaction time and summarize self-reported decision-making strategies. These results are exploratory; there were no specific hypotheses. There was no significant main effect of AI condition on participants' reaction time ($F(2,294) = 2.13, {p} = 0.12, \eta^2 = 0.01$). There was only a main effect of dataset condition ($F(1,294) = 28.52, {p} < 0.001, \eta^2 = 0.09$), where participants took an average of 1600 ms longer in the Census condition than the COMPAS condition (see Figure \ref{fig:rtplot}). This effect was most likely due to the Census dataset having more variables for each instance than the COMPAS dataset, and thus requiring more reading time on each trial. The addition of an explanation did not meaningfully increase reaction time over an AI prediction only.  

Subjective measures, such as self-reported strategies and measures of usability, often diverge from objective measures of human performance ~\cite{andre1995users, nisbett1977telling} such as actual decisions. Participants self-reported varying strategies to make their decisions, yet there was a clear benefit for AI prediction (without and with explanation). In the AI and AI explanation conditions: ${n} = 80$ indicated using the data without mentioning AI, ${n = 39}$ reported using a combination of the data and the AI, and only ${n} = 16$ said they primarily used, trusted, or followed the AI. Despite limited self-reported use of the AI in the two relevant conditions, decision accuracy was higher with AI (Figure \ref{fig:accuracyplot}), strongly associated with AI accuracy (Figure \ref{fig:AIdmacc}), and there was some sensitivity to whether the AI was followed when it was correct versus incorrect (Figure \ref{fig:aiplot2}). 
Nearly 80\% of user comments could be coded; blank and nonsense responses could not be coded.  

\subsection{Discussion}
Our results show providing an AI prediction enhances human decision accuracy, but in opposition to the hypotheses, adding an explanation had no significant impact on decisions and the ability to outperform the AI. Past research has focused on user trust and understanding for explainable AI, rather than objective performance for decision-making. The lack of a significant, practically relevant effect for explainable AI was not due to lack of statistical power or ceiling performance---nearly all participants consistently performed above chance, but well below perfect accuracy. These findings also illustrate the need to compare decision-making with explainable AI to other conditions including no AI and AI prediction without explanation. If we did not have an AI-only (decision aid) condition, a reasonable but flawed inference would have been that explainable AI enhances decisions.

\section{Conclusions and Future Work}

Existing research on explainable AI focuses on the usability, trust, and interpretability of the explanation. In this paper, we fill in the research blank by investigating whether explainable AI can improve human decision-making. We design a behavioral experiment in which each participant recruited using Amazon Mechanical Turk is asked to complete 50 test trials in one of six experimental conditions. Our experiment is conducted on two real datasets to compare human decision with an AI prediction and an AI  with explanation. Our experimental results demonstrate that AI predictions alone can generally improve human decision accuracy, while the advantage of explainable AI is not conclusive. We also show that users tend to follow AI predictions more often when the AI predictions are accurate. In addition, AI with or without explanation can increase the confidence of human users which, on average, was well-calibrated to user decision accuracy.    

In the future, we plan to investigate whether explainable AI can help improve fairness, safety, and ethics by increasing the transparency of AI models. Human decision-making is a key outcome measure, but it is certainly not the only goal for explainable AI. We also plan to explore the difference of distributions in the error space between human decision and AI predictions, especially at decision boundaries. Also, whether human-machine collaboration is feasible through interactions in closed feedback loops. We will also expand our datasets to include other data formats such as images. 



\section*{Broader Impact}

For many important critical decisions, depending on the AI model prediction may not be enough. Furthermore, many recent regulations such as General Data Protection Regulation (GDPR) \cite{euro_reg} allow the human to potentially audit of AI predictions. Therefore, it is critical to understand whether the explanations provided by explainable AI methods improve the overall prediction accuracy and help human decision makers to detect errors. To our knowledge, this is the first work that tries to understand the impact of explainable AI models in improving human decision-making. Our results indicate that although the existence of an AI model may improve human decision-making, the explanations provided may not automatically improve the accuracy.  We believe that our results could help ignite the needed research to explore how to better integrate explanations, AI models, and human operators to have better outcomes compared with AI models or humans alone.

\begin{ack}
The views and conclusions contained in this document are those of the authors and should not be interpreted as representing the official policies, either expressed or implied, of the U.S. Army Combat Capabilities Development Command Army Research Laboratory or the U.S. Government. The U.S. Government is authorized to reproduce and distribute reprints for Government purposes notwithstanding any copyright notation. M.K. and Y.Z. were supported by the Army Research Laboratory/Army Research Office under grant W911NF-17-1-0356, and NSF IIS-1939728. We thank Jason Radford for help implementing the experiment on the Volunteer Science platform and Katelyn Morris for independently coding the comments. We also thank Dan Cassenti for input on the paper, and Jessica Schultheis and Alan Breacher for editing the paper. 
\end{ack}

\bibliographystyle{IEEEtran}
\bibliography{ref}


\end{document}


\title{Supplementary Material for "Does Explainable Artificial Intelligence Improve Human Decision-Making?"}

\maketitle
\section{Age Distribution}
\begin{figure}[h]
    \centering
    \includegraphics[width=0.5\linewidth]{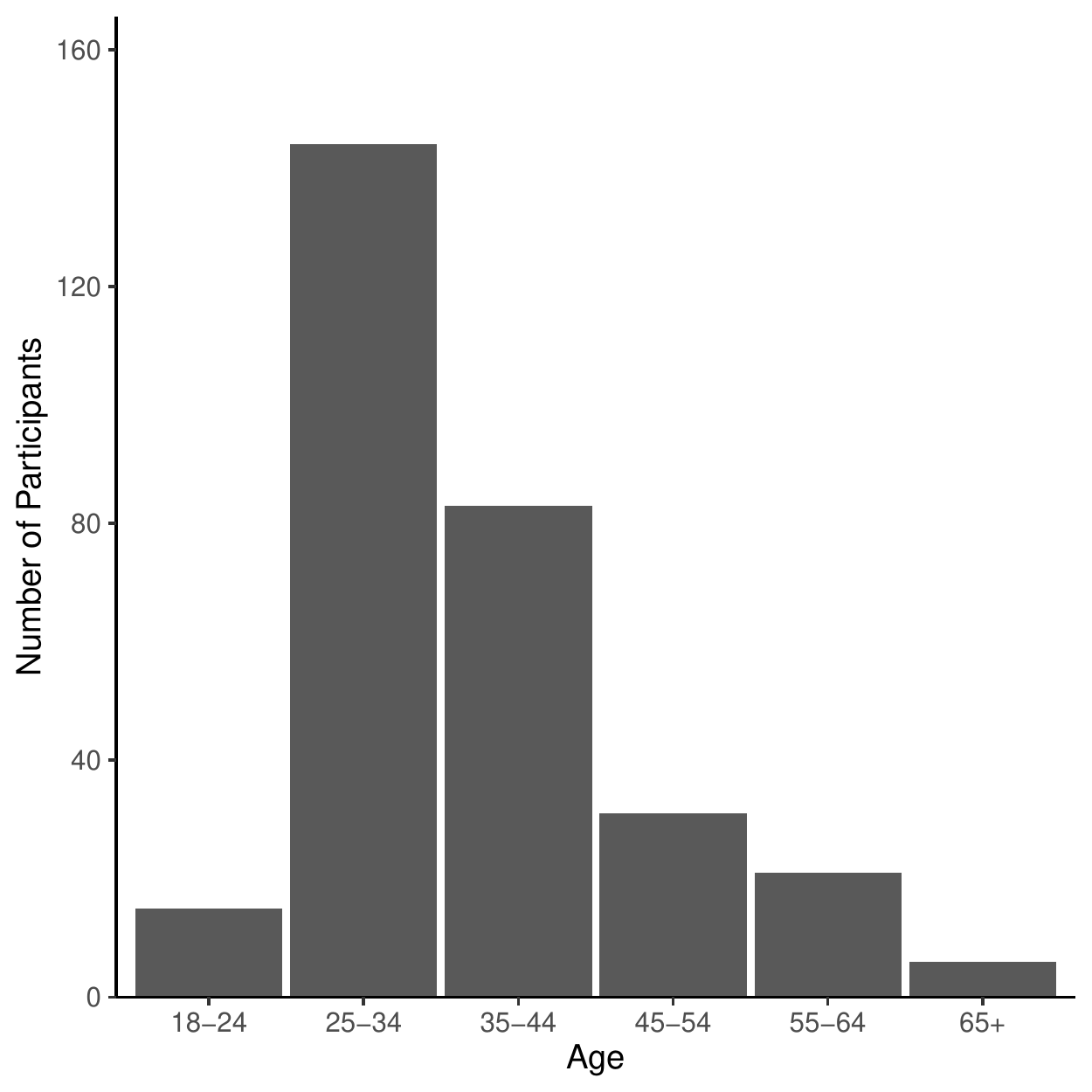}
    \caption{
        Age distribution of participants.
        \label{fig:ageplot}
    }
\end{figure}

See Figure \ref{fig:ageplot} for the age distribution of participants. Most participants were in the 25-34 age group. 

\section{Self-Reported Decision Strategies} 
We performed an exploratory analysis on the self-reported decision strategies, qualitatively described by users in their comments. We coded strategies using six categories---see the first column in Table~\ref{tab:cond}. First, comments were independently coded by two raters: the third author and a research assistant. Inter-rater reliability was good, $\kappa = 0.70, {p} < 0.001$. Second, we reached consensus to resolve all discrepancies and finalize categorical coding of decision strategies.

The majority of participants commented that they primarily relied on the data to make decisions, especially in the control condition: ${n} = 74$. The dominant reported strategy in the two AI conditions was primarily using the data to make decisions without any mention of AI: ${n} = 80$. Primary use of AI (${n} = 16$) was surprisingly low. Although partial use of AI, including the possibility of explanations in the explainable AI condition, was higher (${n} = 39$), it was still limited compared to the more widely used strategy of making decisions using the data, whereas the number of participants using each strategy was generally comparable between the two datasets, see Table~\ref{tab:dataset}. There were minor differences in more frequent use of the AI and data together for COMPAS over the Census dataset. This was despite the higher mean AI accuracy for the Census income dataset over COMPAS.

\begin{figure}[b]
    \centering
    \includegraphics[width=0.8\linewidth]{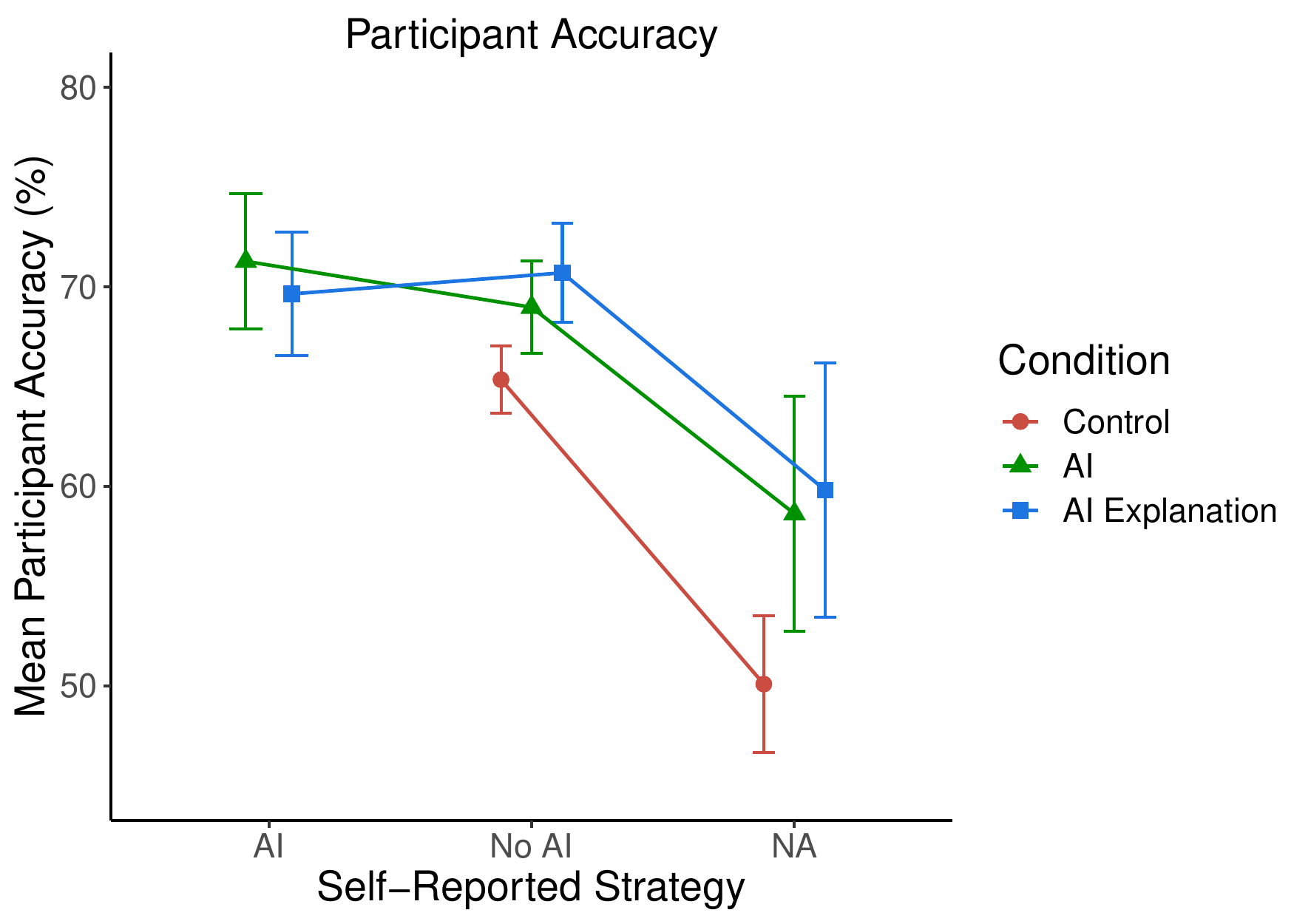}
    \caption{
        Participant accuracy by self-reported strategy and AI condition. Error bars are 95\% confidence intervals. 
        \label{fig:accstratplot}
    }
\end{figure}

We compared the accuracy of participants who self-reported different strategies, grouping them into those who mentioned using AI (in the AI and AI Explanation conditions; Primarily AI and AI and data), those who did not mention using AI (Data and gut, Primarily gut, and Primarily data), and those who provided blank or nonsense strategies (Could not be coded, labelled as NA). As shown in Figure \ref{fig:accstratplot}, mean participant accuracy was very similar for those who mentioned using the AI versus those who did not. In the future, it may be useful to compare the types of decision errors for AI versus AI with explanation as well as errors based on self-reported strategies. Different types of errors could produce similar decisional accuracy, if the overall error rate is similar. However, participants whose self-reported strategies were blank or did not make sense did show a lower mean accuracy on the task, perhaps indicating a lower level of effort or engagement with the task overall with accuracy nearing or even at chance performance (50\%). The number of users with no self-reported strategy was comparable across all six conditions.   

\begin{center}
 \begin{tabular}{||c c c c||} 
 \hline
 Strategy & Control & AI & AI Explanation \\ [0.5ex] 
 \hline\hline
 Primarily AI & -- & 7 & 9 \\ 
 \hline
 AI and data  & -- & 14 & 25\\
 \hline
 Data and gut & 5 & 2 & 1   \\
 \hline
 Primarily gut & 2 & 6 & 6 \\
 \hline
 Primarily data & 74 & 46 & 34 \\ 
  \hline
 Could not be coded & 19 & 25 & 25 \\ [1ex]
  \hline
\end{tabular}
\label{tab:cond}
\end{center}

\begin{center}
 \begin{tabular}{||c c c||} 
 \hline
 Strategy & Census & COMPAS \\ [0.5ex] 
 \hline\hline
 Primarily AI & 9 & 7\\ 
 \hline
 AI and data  & 16 & 23\\
 \hline
 Data and gut & 4 & 4 \\
 \hline
 Primarily gut & 3 & 11\\
 \hline
 Primarily data & 83 & 71 \\ 
  \hline
 Could not be coded & 35 & 34\\ [1ex]
  \hline
\end{tabular}
\label{tab:dataset}
\end{center}

\section{Speed-Accuracy Tradeoff}

\begin{figure}[h]
    \centering
    \includegraphics[width=0.8\linewidth]{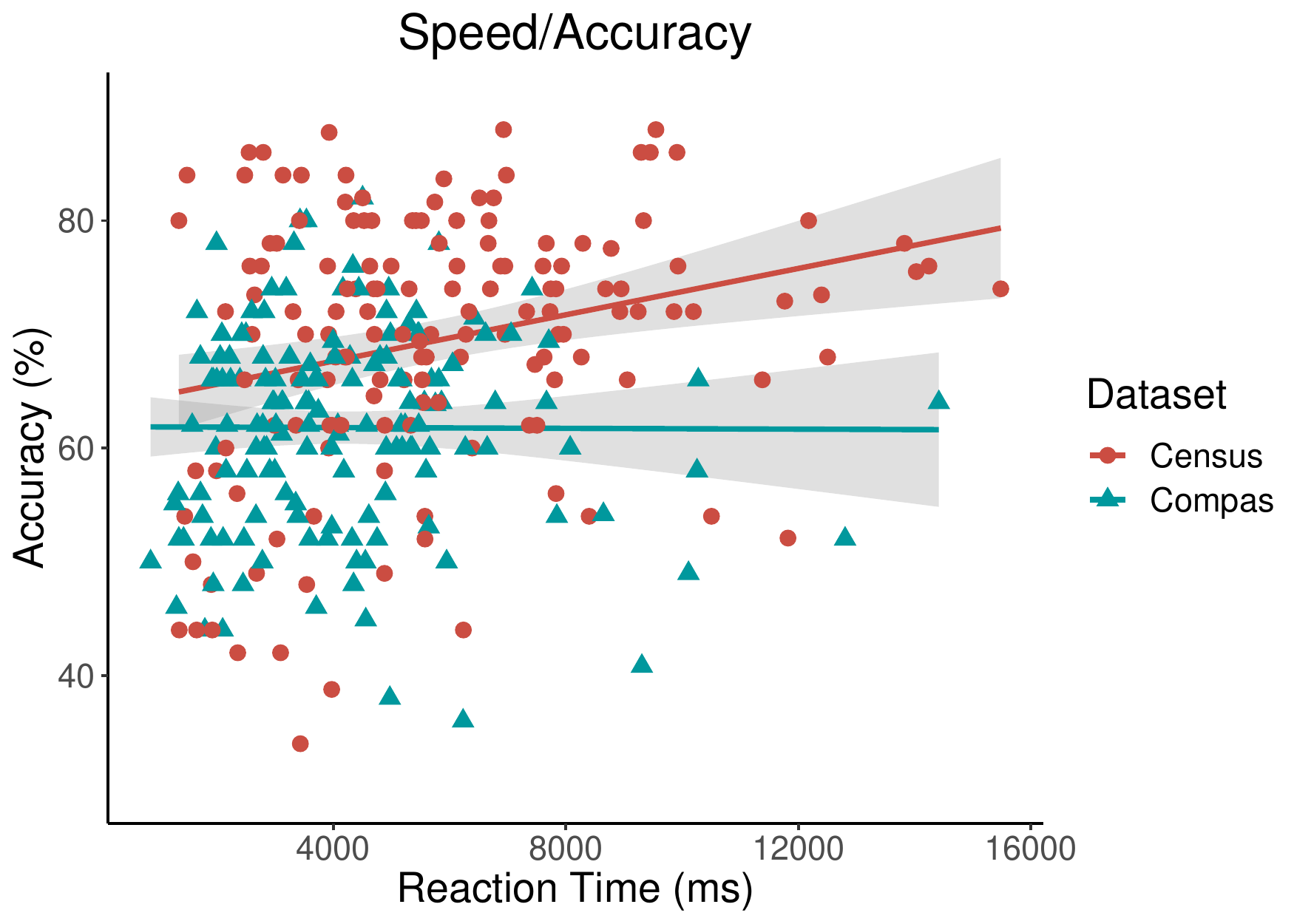}
    \caption{
        Speed-accuracy tradeoff generated by participants in each dataset condition. Shaded areas indicate 95\% confidence intervals. 
        \label{fig:sato}
    }
\end{figure}

We used linear regression to explore the tradeoff between speed and accuracy across participants---that is, whether slower participants tended to be more accurate and vice versa. We found overall a small but significant speed-accuracy tradeoff ($\beta = 1.01, {p} < 0.001, \eta^2 = 0.08$), where each additional second of reaction time predicted a 1.01\% increase in accuracy. As shown in Figure \ref{fig:sato}, there was a small interaction between the dataset condition and reaction time ($F(1,288) = 6.65, {p} < 0.02, \eta^2 = 0.02$), indicating that the speed-accuracy tradeoff was present for participants in the Census dataset condition ($\beta = 1.02$), but it was near zero for those in the COMPAS dataset condition ($\beta = -0.02$). We did not find a significant effect of AI condition on the speed-accuracy tradeoff. 

